\documentclass[conference]{IEEEtran}
\IEEEoverridecommandlockouts
\usepackage{cite}
\usepackage{amsmath,amssymb,amsfonts}
\usepackage{algorithmic}
\usepackage{graphicx}
\usepackage{textcomp}
\usepackage{xcolor}
\usepackage{algorithm}
\usepackage{algorithmic}
\usepackage{booktabs}
\usepackage{xcolor}
\usepackage{url}
\usepackage{latexsym}
\usepackage{amssymb}
\usepackage{amsmath}
\usepackage{amsthm}
\usepackage{booktabs}
\usepackage{enumitem}
\usepackage{graphicx}
\usepackage{color}
\usepackage{url}
\usepackage{hyperref}

\usepackage{multirow}

\title{Learning Distinguishable Representations in Deep Q-Networks for Linear Transfer}
\author{\IEEEauthorblockN{Sooraj Sathish\thanks{This work has been accepted at ICTAI 2025 as a full paper.}}
\IEEEauthorblockA{\textit{IIIT Bangalore}, India \\
sooraj.sathish@iiitb.ac.in}
\and
\IEEEauthorblockN{Keshav Goyal}
\IEEEauthorblockA{
\textit{IIIT Bangalore},
India \\
keshav.goyal@iiitb.ac.in}
\and
\IEEEauthorblockN{Raghuram Bharadwaj Diddigi}
\IEEEauthorblockA{
\textit{IIIT Bangalore},
India \\
raghuram.bharadwaj@iiitb.ac.in}
\thanks{Dr. Raghuram Bharadwaj is supported by the Anusandhan National Research Foundation (ANRF) under the Prime Minister Early Career Research Grant ANRF/ECRG/2024/005235/ENS}
}
\begin{document}

\maketitle

\begin{abstract}
Deep Reinforcement Learning (RL) has demonstrated success in solving complex sequential decision-making problems by integrating neural networks with the RL framework. However, training deep RL models poses several challenges, such as the need for extensive hyperparameter tuning and high computational costs. Transfer learning has emerged as a promising strategy to address these challenges by enabling the reuse of knowledge from previously learned tasks for new, related tasks. This avoids the need for retraining models entirely from scratch.
A commonly used approach for transfer learning in RL is to leverage the internal representations learned by the neural network during training. Specifically, the activations from the last hidden layer can be viewed as refined state representations that encapsulate the essential features of the input. In this work, we investigate whether these representations can be used as input for training simpler models, such as linear function approximators, on new tasks. We observe that the representations learned by standard deep RL models can be highly correlated, which limits their effectiveness when used with linear function approximation.
To mitigate this problem, we propose a novel deep Q-learning approach that introduces a regularization term to reduce positive correlations between feature representation of states. By leveraging these reduced correlated features, we enable more effective use of linear function approximation in transfer learning. Through experiments and ablation studies on standard RL benchmarks and MinAtar games, we demonstrate the efficacy of our approach in improving transfer learning performance and thereby reducing computational overhead.
\end{abstract}




         








\section{Introduction}

Reinforcement Learning (RL) \cite{bertsekas1995dynamic, suttonandbarto} is a popular paradigm for solving sequential decision-making problems under uncertainty. When combined with deep neural networks, RL has achieved success in various domains, such as games \cite{silver2016mastering,mnih2013playing,mnih2015human}, robotics \cite{ibarz2021train,gu2017deep}, autonomous driving \cite{kiran2021deep,sallab2017deep}, and human alignment of large language models \cite{ouyang2022training}. However, this integration introduces challenges, including increased computational demands and the need for careful hyperparameter tuning. Moreover, RL algorithms often require significant GPU resources, adding to the overall computational complexity. 
\begin{figure}
    \centering
    \includegraphics[scale = 0.5]{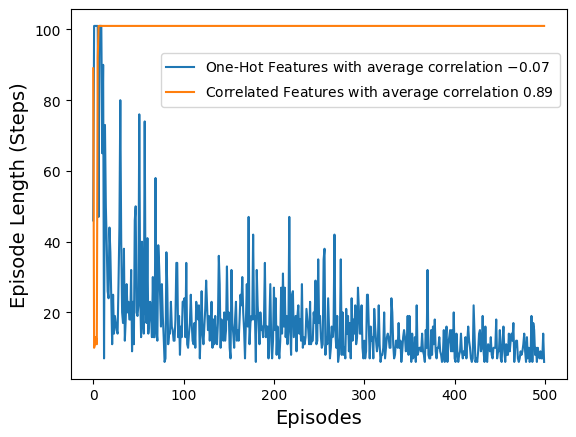}
    \caption{Performance of linear function approximation Q-learning on a $4\times4$ grid world. Feature sets with high positive correlation result in a bad policy, with longer episode lengths (maximum length capped at $100$). In contrast, feature sets with lower positive correlation enable faster learning and convergence to shorter episode lengths. Both set of features represent the same current position.}
    \label{introfig}
\end{figure}
Given these challenges, transfer learning has become increasingly important in RL \cite{transfer_survey}. The ability to apply knowledge from previously trained models to solve related tasks is more efficient than training new models from scratch. For instance, consider a household robot trained via deep RL to move an object from a source to a designated target. If the target location changes slightly, the existing policy will no longer be effective. Retraining the model from scratch for such a minor variation is inefficient; instead, it would be more practical to leverage the knowledge from the pre-trained model. Additionally, it is not desirable to use complex deep RL algorithms for training the new task, instead one prefers to train a simpler and faster model. 

A widely adopted method for transfer learning in RL involves training a deep RL neural network on an initial task and then utilizing its hidden layers for a new, related task. In this work, we focus on the Deep Q-Learning (DQN) \cite{mnih2013playing}, a popular deep RL algorithm. The DQN model takes the state of the environment as input and predicts Q-values for all possible actions, representing the expected cumulative reward for each action. An interesting interpretation of DQN is to view it as the network that transforms the input state into a more meaningful and compact representation before combining this representation linearly to generate the Q-values \cite{DRL_decor}. Specifically, the output of the final hidden layer can be viewed as a refined state representation. We propose leveraging these refined state representations from a trained DQN model and using them as state features to train a linear function approximation Q-learning for the new task, thereby facilitating more efficient and simpler transfer learning.

However, transferring these improved state representations is not without its challenges. A key requirement for effective linear function approximation is that the state features must be sufficiently distinguishable to ensure faster learning. Unfortunately, deep neural networks are prone to overfitting \cite{farebrother2018generalization}, which can lead to highly correlated state representations. We observe that, in standard DQN, the outputs from the final hidden layer, which represent the state features, can exhibit strong positive correlations. This becomes a significant challenge for the linear function approximator, as it becomes difficult to learn optimal weights for the new task. For instance, in Figure \ref{introfig}, we demonstrate the impact of highly correlated features in a grid world environment, where the objective is to reach the target state from an initial state in fewer steps. The linear function approximator struggles to learn even suboptimal behaviour when the state features are highly correlated (average correlation across the states is $0.89$), whereas features with less correlation (average correlation of $-0.07$) allow for learning the optimal path more effectively.

To address this issue, we introduce a regularization term to the DQN loss function during training that penalizes positive correlations between state features. This modification ensures that the trained DQN not only performs well on the original task but also produces less positively correlated state features. While the goal of function approximation is to generalize effectively across states, this regularization term ensures that positive correlations are preserved only where necessary. Consequently, the resulting state features are more distinguishable and can be leveraged for transfer to new tasks.


To this end, we first present an interpretation of DQN as a network that transforms raw states into more refined state representations, which are then linearly combined with weights to produce Q-values. This viewpoint allows us to introduce a regularization term into the DQN loss function to reduce feature correlation. To optimize this new loss function, we propose an algorithm to effectively train with this modified loss function. Using the learned state features, we then apply linear function approximation (LFA) Q-learning. Our results demonstrate that LFA Q-learning performs effectively in complex environments, such as Lunar Lander and MinAtar, which were previously considered unlearnable with linear methods alone. Additionally, we show that the model adapts well to new environments, even with slight modifications in reward structures, thus achieving transfer learning with significantly reduced computational overhead.
\section{Related Works}
In this section, we review works from the literature, categorizing them into three key related areas: representation learning and feature decorrelation, transfer learning, and linear function approximation.

\textit{Decorrelation and Representation Learning}. 
Sparse Coding \cite{suttonandbarto, suttonsparse} is an effective but computationally demanding way of designing feature representations in RL. 
\cite{sparse_rep} has shown that sparse representations can help address environments that are too complex for direct neural network approaches. 
In the context of Deep Reinforcement Learning, \cite{DRL_decor} propose a method for efficient representation learning using DQN features. They observe that DQN features are inherently correlated and demonstrate that adding a regularization term to decorrelate latent features significantly improves model performance in complex environments, such as Atari. This idea is further extended in \cite{
mavrin2019efficient} with a theoretical justification for the decorrelation term. 

It is \textbf{important to note} that in \cite{DRL_decor}, the correlation is computed between feature components of a state, while in our work, we address correlations across features of different states. This distinction is critical, as our approach is specifically motivated to facilitate efficient transfer learning, whereas \cite{DRL_decor} is aimed at faster learning of DQN. 
\cite{URL} also highlights how decorrelating latent space features prevents representational collapse, a key issue in unsupervised representation learning (URL). Their approach improves sample efficiency by increasing the dimensionality of the latent manifold, which benefits URL methods in reinforcement learning tasks.

\textit{Transfer Learning}. Transfer learning in RL has evolved through various approaches designed to accelerate skill acquisition across tasks. In \cite{invariance}, invariant feature spaces are employed to facilitate knowledge transfer between agents with different morphologies, such as robotic arms. Similarly, DARLA \cite{darla} achieves zero-shot transfer \cite{zero_shot2} by learning disentangled visual representations, enabling policy transfer across domains without requiring access to target domain data. \cite{farebrother2018generalization} studies the impact of dropouts and $L_2$ regularization on generalization and transfer capabilities of DQN. \cite{transfer_survey, transfer_survey2} provides a comprehensive review of various transfer learning techniques pertaining to RL. 


\textit{Linear Function Approximation}. Linear models have been investigated as a more practical alternative to complex deep RL approaches, particularly in environments with high-dimensional inputs. In \cite{linear1}, it is shown that model-free linear approximation methods obtain the same value as algorithms that utilise knowledge and dynamics of the environment (model-based algorithms). This relationship provides insights into feature selection by decomposing Bellman error into reward and feature errors, thus offering a clearer understanding of feature-selection behaviour in RL. In \cite{linear2}, the authors evaluate linear models in the Arcade Learning Environment (ALE). By leveraging representational biases from DQN, such as the use of convolutional neural networks (CNNs), object detection, and non-Markovian features, they achieve competitive performance on the ALE benchmark.

However, to the best of our knowledge, none of the existing works in the literature have explored explicitly training DQN with reduced positive correlation across state representations to enable more efficient transfer learning. The main contributions of our work are as follows:

\begin{enumerate}
\item We propose a new DQN algorithm that employs a regularization term to penalize positive correlations, leading to reduced positively correlated state representations. 
\item By leveraging these learned state representations, we demonstrate the effectiveness of Linear Function Approximation in solving complex environments. Additionally, we show that this approach enables efficient transfer learning on challenging benchmark RL environments. 
\item We perform thorough ablation studies analyzing various components of the proposed system. 
\end{enumerate} 

\section{Background}


 

A Markov Decision Process (MDP) is a widely used mathematical framework for solving sequential decision-making problems . An MDP is characterized by the tuple $(S, A,T,\gamma, D,R)$ \cite{puterman2014markov}, where $S$ represents the set of states, and $A$ denotes the set of actions. The state transition function, $T: S \times A \times S \to [0,1]$, governs the dynamics of the environment by defining the probability of transitioning from one state to another after taking a particular action. $D$ represents the initial state distribution, and $\gamma \in [0,1)$ is the discount factor that controls the importance of future rewards. The reward function, $r: S \times A \to \mathbb{R}$, provides feedback based on the action taken in a given state. The goal of a reinforcement learning (RL) agent in an infinite horizon setup \cite{bertsekas1996neuro} is to learn a policy $\pi^*: S \to A$ that maximizes the expected cumulative reward, which is defined as: \begin{align} \pi^* = \arg \max_{\pi \in \Pi}\mathbb{E} \left[\sum_{t=0}^{\infty }\gamma^t r(s_t,\pi(s_t)) \mid s_0 \sim D \right], \end{align} where the expectation is taken over all states encountered at time steps $t = 0, \ldots, \infty$ and $\Pi$ is the set of all possible policies.

When the transition function of the environment is unknown, model-free reinforcement learning (RL) algorithms \cite{suttonandbarto} are employed to compute the optimal policy. One widely used model-free RL algorithm is Q-learning \cite{watkins1992q}. Q-learning maintains a table of Q-values, $Q(s, a)$, which are updated iteratively based on the observed state, action, reward, and next state sample. Specifically, given the tuple $(s_t,a_t, r_{t+1}, s_{t+1})$, the Q-values at time $t$ are updated as follows:
\begin{align} Q(s_t, a_t) \leftarrow Q(s_t, a_t) + \alpha \Big[ r_{t+1} + &\gamma \max_{a'} Q(s_{t+1}, a') \\ & - Q(s_t, a_t) \Big],\nonumber \end{align}
where $\alpha$ is the learning rate, and the Q-values for all state-action pairs are initialized randomly. If each state-action pair is visited sufficiently often, the Q-values converge to the optimal Q-values, $Q^*(s,a)$, as $t \to \infty$ \cite{watkins1992q}. Here, $Q^*(s,a)$ represents the optimal expected cumulative reward when action $a$ is taken in state $s$. The optimal policy, $\pi^*$, can then be derived as:
\begin{align} \pi^*(s) = \arg \max_{a \in A} Q^*(s,a), \quad \forall s \in S. 
\end{align}

When dealing with large or continuous state spaces, the traditional Q-learning algorithm becomes impractical, as storing all possible $Q(s,a)$ values in a tabular format is infeasible. In such cases, Linear Function Approximation (LFA) Q-learning is used, where the Q-values are approximated as a linear combination of state-action features: $Q(s,a) = w^\top \phi(s,a)$, where $\phi(s,a)$ represents the feature vector corresponding to the state $s$ and action $a$.
Given the tuple $(s_t,a_t, r_{t+1}, s_{t+1})$ at time $t$, the Q-learning update for linear function approximation modifies the parameters $w$, rather than individual Q-values, as follows:
\begin{align} 
w \leftarrow w + \alpha \Big[ r_{t+1} + \gamma &\max_{a'} w^\top \phi(s_{t+1}, a') \\ & - w^\top \phi(s_t, a_t)\Big] \phi(s_t, a_t). \nonumber
\end{align}

Deep Q-Networks (DQN) \cite{dqn} eliminate the need for hand-crafted state-action features by using a neural network to directly approximate Q-values. The parameters of the neural network, denoted by $\theta$, are updated by minimizing the following loss function:
\begin{equation} L(\theta) = \mathbb{E} \left[ \left( y_t - Q(s_t, a_t; \theta) \right)^2 \right], \end{equation}
where
\begin{equation} y_t = r_{t+1} + \gamma \max_{a'} Q(s_{t+1}, a'; \theta^-), \end{equation}
and $\theta^-$ are the parameters of the target network. A key heuristic introduced in \cite{dqn} is the use of two separate networks: the policy network, which learns the action-value function, and the target network, which provides stable targets for the loss function. The target network’s parameters are updated less frequently by copying the policy network's weights periodically.
DQN also employs Experience Replay \cite{experience_replay}, where transitions $(s_t,a_t,r_{t+1},s_{t+1})$ are stored in a replay buffer. Instead of updating after every transition, random mini-batches are sampled from this buffer for training. This technique helps to decorrelate the training data and stabilizes the learning process.

It is important to note that while DQN performs well in various settings, it is highly sensitive to hyperparameters and the initial random weights. Additionally, the learned weights can become overfitted to the specific task, making it difficult to transfer them effectively to related tasks and achieve faster learning. Furthermore, training DQN models requires significant computational resources, such as high-performance GPUs, and is time-consuming. Our objective is to reduce overfitting in the network in a manner that facilitates efficient transfer to simpler frameworks, such as linear function approximation, for solving related tasks.


\section{Proposed Algorithm}


In this section, we first begin by presenting a perspective on DQN that allows us to impose constraints on state representations (Section \ref{viewpoint}). We then introduce a novel DQN loss function designed to address these constraints, followed by the proposal of a new DQN algorithm (Section \ref{decorrelation}). Lastly, we discuss how these learned state representations can be effectively transferred to a linear function approximation setting for solving related tasks (Section \ref{lfa}).


\subsection{Alternative Viewpoint of the DQN}\label{viewpoint}
As discussed in the previous section, the goal of DQN is to accurately estimate the Q-values for the optimal policy. The network takes a raw state $s$ as input and outputs $Q(s,a)$ for all actions $a \in A$. An interesting interpretation of DQN is to view it as a function that first transforms the raw state into a richer representation, which is then linearly combined to compute the Q-values \cite{DRL_decor}. Let $\theta = (\theta_1, w)$ represent the parameters of the DQN neural network, where $\theta_1$ corresponds to the parameters up to the final hidden layer, and $w(a)$ is the weight of the final layer for action $a$. In this case, the Q-value for any action $a$ and state $s$ can be expressed as:
\begin{equation} Q(s,a) = \phi_{\theta_1}(s)^\top w(a), \end{equation}
where $\phi(s)$ \footnote{We drop the subscript $\theta_1$ for notational convenience and easier readability} is the output of the final hidden layer.
Thus, the objective in DQN is:
\begin{equation} \label{vdqn}
\min_{\theta_1, w} \mathbb{E} \left[ \left( y - \phi(s)^\top w(a) \right)^2 \right], \end{equation}
where
\begin{equation} y = r + \gamma \max_{a'} \phi(s')^\top w(a'). \end{equation}

The principle behind the DQN is to generalize learning across different states, tuning the network’s weights in a way that minimizes the loss function defined by \eqref{vdqn}. However, during training, the network may overgeneralize, resulting in state representations $\phi(s), ~ s \in S$ that are very similar across different states. While this might not significantly impact the performance of DQN itself, since the weights $w(a)$ are being adjusted accordingly, this overgeneralization creates a challenge when transferring these features to simpler methods like linear function approximation (LFA). LFA may struggle to learn or recover the appropriate weights from these overly similar representations. 

By adopting this viewpoint, we decouple the state representations and the weights, allowing us to impose constraints specifically on the state representations $\phi(s)$. This flexibility is crucial for ensuring that the learned features remain distinguishable, facilitating more effective transfer learning to simpler algorithms like LFA. Thus, our goal is to develop a modified DQN algorithm that not only computes the optimal policy but also generates more distinguishable state representations, making the transfer process smoother and more efficient.



\subsection{Reducing Positive Correlation between the states}\label{decorrelation}
 

To improve the distinguishability across state representations and facilitate transfer learning, we introduce a regularization term into the DQN loss function that penalizes high positive correlation between the representations of different states. The modified DQN loss function is defined as:
\begin{align} \label{newfor} L_1(\theta_1,w) + \lambda L_2(\theta_1), \end{align}
where
\begin{align} L_1(\theta_1,w) = \mathbb{E} \left[ \left( y - \phi(s)^\top w(a) \right)^2 \right], \end{align}
is the standard DQN loss function, and
\begin{align}
    L_2(\theta_1) = \displaystyle \mathbb{E}_{s_a, s_b} \left[ \textbf{corr}(\phi(s_a), \phi(s_b))\right],
\end{align}
where $\textbf{corr}(\phi(s_a), \phi(s_b))$ is the Pearson correlation coefficient between the feature vectors $\phi(s_a)$ and $\phi(s_b)$ for two independently sampled states $s_a$ and $s_b$. If $\phi(s_a) \in \mathbb{R}^n$, the Pearson correlation coefficient is given by:
\begin{align}
    \frac{\sum_{i=1}^{n} \left( \phi_i(s_a) - \bar{\phi}(s_a) \right) \left( \phi_i(s_b) - \bar{\phi}(s_b) \right)}{\sqrt{\sum_{i=1}^{n} \left( \phi_i(s_a) - \bar{\phi}(s_a) \right)^2} \sqrt{\sum_{i=1}^{n} \left( \phi_i(s_b) - \bar{\phi}(s_b) \right)^2}},
\end{align}
where $\phi_i(s_a)$ is the $i$-th feature of state $s_a$, and $\bar{\phi}(s_a)$ is the mean of the features of state $s_a$. $0 \leq \lambda \leq \lambda_{max}$ is the regularization parameter. 

\begin{algorithm}[ht]
\caption{Distinguishable States DQN (DS-DQN)}
\label{UC-DQN}
\begin{algorithmic}[1]  
    \STATE \textbf{Initialize:} Replay buffer $R$ to capacity $N$, weights $\theta = (\theta_1, w)$, target network weights $\theta^{-}$, learning rates $\alpha, \beta$, regularization coefficient $\lambda$ initialized to $0$, regularization increment $\Delta \lambda$, maximum regularization coefficient $\lambda_{\text{max}}$
    \STATE \textbf{Set:} Target network parameters $\theta^{-} \leftarrow \theta_1$
    
    \FOR{each episode}
        \STATE Initialize state $s$
        
        \FOR{each step of the episode}
            \STATE With probability $\epsilon$, select a random action $a$, otherwise select $a = \arg\max_{a} Q(s,a;\theta)$
            \STATE Execute action $a$ and observe reward $r$ and next state $s'$
            \STATE Store transition $(s,a,s',r)$ in replay buffer $R$
            
            \STATE Sample a random mini-batch of transitions $(s_j, a_j, r_j, s_j')_{j=1}^{m}$ from replay buffer $R$
            
            \STATE Compute the target for each mini-batch sample:
            \[
            y_j = \begin{cases} 
            r_j & \text{if } s_j' \text{ is terminal} \\
            r_j + \gamma \max_{a'} Q(s_j', a'; \theta^{-}) & \text{otherwise}
            \end{cases}
            \]

            \STATE Compute the DS-DQN loss $L_1(\phi, w)$:
            \[
            L_1(\phi, w) = \frac{1}{m}\sum_{j=1}^{m} \left[ (y_j - Q(s_j, a_j; \theta_1))^2 \right]
            \]

            \STATE Select $l$ pairs of $\{(s_a^k, s_b^k)\}_{k=1}^{l}$, sampled independently, and compute regularization loss $L_2(\phi)$:
            \[
            L_2(\phi) =  \frac{1}{l}\sum_{k=1}^{l} \text{corr}(\phi(s_a^k), \phi(s_b^k))
            \]

            
            \STATE Update the network parameters:
            \[
            \theta_1 \leftarrow \theta_1 - \alpha \left( \nabla_{\theta_1} L_1(\phi,w) + \lambda \nabla_{\theta_1} L_2(\phi) \right)
            \]
            \[
            w \leftarrow w - \beta \nabla_{w} L_1(\phi,w)
            \]
        \ENDFOR
        \STATE $\lambda = \min(\lambda + \Delta \lambda, \lambda_{\text{max}})$
        \STATE Every $C$ steps, update the target network:
            \[
            \theta^{-} \leftarrow \theta_1
            \]
    \ENDFOR
\end{algorithmic}
\end{algorithm}

Our goal is to produce distinguishable state representations that facilitate efficient transfer to linear function approximation (LFA). This is achieved through the interaction between the losses $L_1$ and $L_2$. While minimizing $L_1$ can lead to the network making the representations of different states similar, $L_2$ penalizes high correlation between these representations. This interplay ensures that the state representations are learned in a way that not only optimizes the original task but also makes them sufficiently distinguishable, enabling effective transfer learning.
The complete pseudo-code of our proposed algorithm, ``Distinguishable States DQN (DS-DQN)'', is presented in Algorithm \ref{UC-DQN}. 
After training the DS-DQN algorithm, the learned state representations $\phi(s)$ can be effectively leveraged for transfer to the LFA framework.


\begin{algorithm}
\caption{LFA Q-learning Algorithm with Enhanced State Representation}
\label{LFAalgo}
\begin{algorithmic}[1]
    \STATE \textbf{Input:} State features $\phi(s)$ from Algorithm \ref{UC-DQN}.
    \STATE \textbf{Initialize:} Weights $w(a), ~ \forall a$, learning rate $\alpha$
    \FOR{each episode}
        \STATE Initialize state $s$
        
        \FOR{each step of the episode}
            \STATE Select action $a$ using $\epsilon$-greedy policy based on $w(a)$ and enhanced state representation $\phi(s)$
            
            \STATE Execute action $a$ and observe reward $r$, next state $s'$
            
            \STATE Compute the TD-target $y$:
            \[
            y = r + \gamma \max_{a'} \left[ \phi(s')^\top w(a') \right]
            \]
            
            \STATE Compute the TD-error $L$:
            \[
            \delta = y - \phi(s)^\top w(a)
            \]
            
            \STATE Update the weights $w(a)$ for action $a$:
            \[
            w(a) \leftarrow w(a) + \alpha \delta \phi(s)
            \]
            
            \STATE Set $s \leftarrow s'$
        \ENDFOR
        
    \ENDFOR
\end{algorithmic}
\end{algorithm}
\subsection{Transfer to Linear Function Approximation}\label{lfa}
We now apply Linear Function Approximation (LFA) using the enhanced state representations $\phi(s)$ obtained from Algorithm \ref{UC-DQN}. For a given new task, our objective is to estimate the optimal weights $w^*(a)$ for all $a \in A$ such that the optimal Q-values can be expressed as:
\begin{align} Q^*(s,a) = \phi(s)^\top w^*(a). \end{align}
The optimal weights are learned by running the following iterative update on the sample $(s_t,a_t,r_{t+1},s_{t+1})$ obtained at each time step $t$:
\begin{align} 
w(a_t) \leftarrow w(a_t) + \alpha [ r_{t+1} + &\gamma \max_{a'} w(a')^\top \phi(s_{t+1}) \\ \nonumber &- w(a_t)^\top \phi(s_t)] \phi(s_t).
\end{align}
The full pseudo-code for the transfer LFA Q-learning is presented in Algorithm \ref{LFAalgo}.

In this way, we enable efficient transfer from a Deep Q-Network to LFA. With its simplicity and extremely low computational cost, LFA is attractive for scenarios with limited resources. Importantly, LFA is very hard to train directly on complex tasks. Therefore, the idea is to first train our proposed DS-DQN (Algorithm \ref{UC-DQN}) on one task and then use the simpler, more computationally efficient LFA (Algorithm \ref{LFAalgo}) to solve other tasks by leveraging the state representations learned from DS-DQN. Unlike standard DQN, our proposed DS-DQN ensures distinguishable state representations, allowing LFA to compute the optimal weights more effectively. We empirically demonstrate the benefits of this approach in the next section.
\section{Experiments and Results}
In this section, we present the results of our proposed algorithm on five reinforcement learning (RL) tasks: Mountain Car \cite{mountaincar}, Lunar Lander \cite{lunarlander}, and three MinAtar games \cite{young2019minatar}: Breakout, Space Invaders, and Asterix. MinAtar is a simplified version of Atari games that reduces the representational complexity while emphasizing behavioural challenges. Additionally, while standard Atari games are deterministic, MinAtar introduces stochasticity to the environment, making the learning process more challenging. In the first subsection, we describe the experimental setup, followed by a detailed discussion of the results. Lastly, we perform ablation studies to further analyze the algorithm and report our observations.
\subsection{Experimental Setup}
\begin{figure*}[ht]
    \centering
    \includegraphics[width=0.6\textwidth]{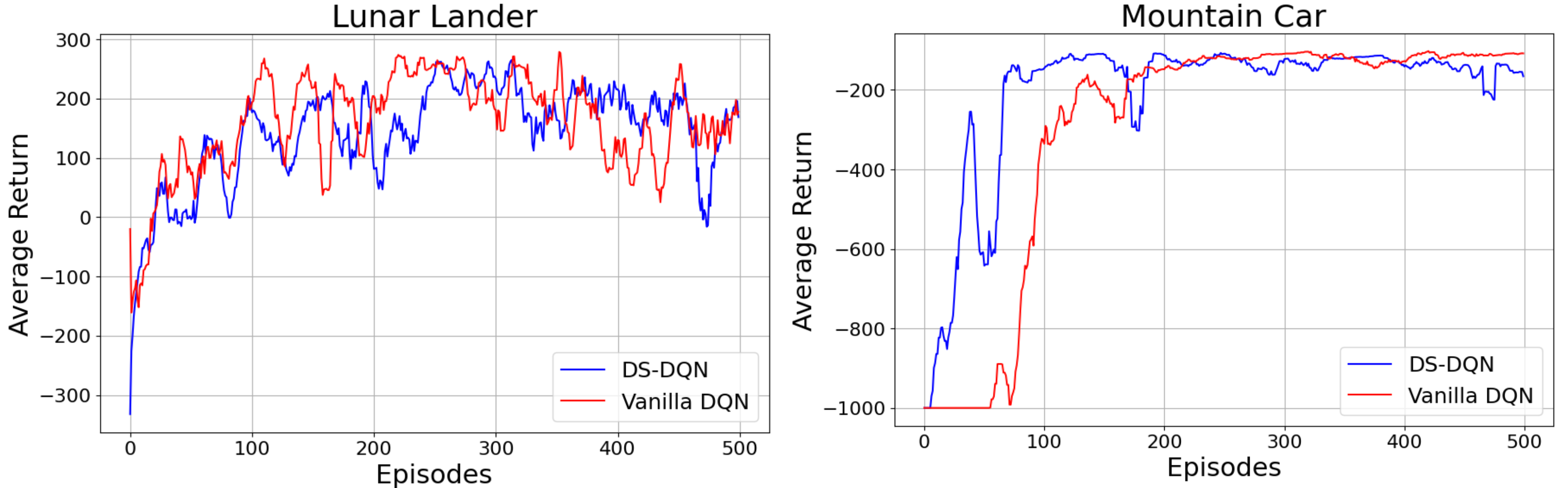} 
    \caption{Average return in last 10 episodes when trained on the Lunar Lander and Mountain Car environments}
    \label{rew_episode}
\end{figure*}

\begin{figure*}[ht]
    \centering
    \includegraphics[width=0.75\textwidth]{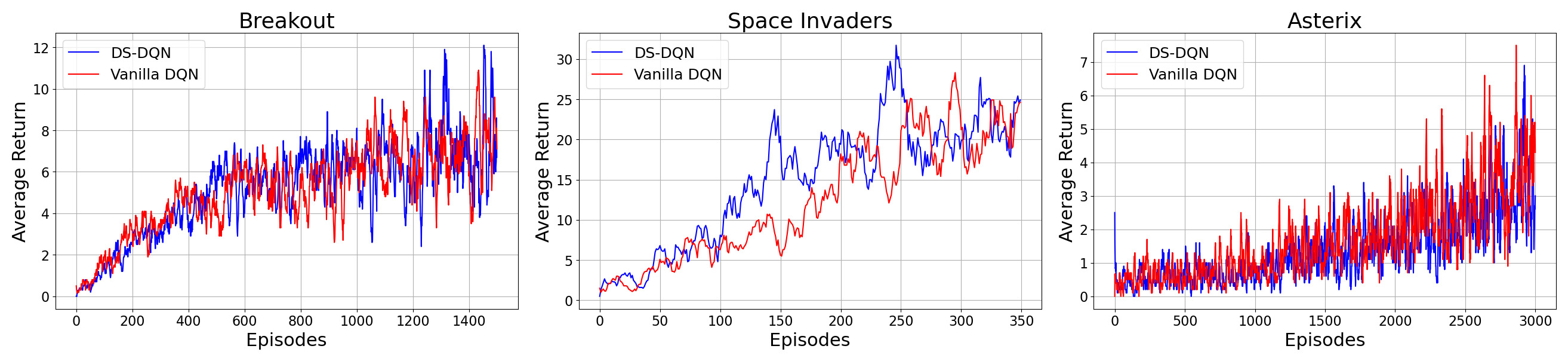} 
    \caption{Average return in last 10 episodes when trained on the MinAtar environments}
    \label{rew_episode_2}
\end{figure*}
\begin{table*}[ht]
\caption{Transfer Performance of LFA Q-learning on state representations from DS-DQN and Vanilla DQN}
\label{main_table}
\centering
\begin{tabular}{|c|c|c|c|c|c|c|c|c|c|}
\hline
Environment & Algorithm & Maximum Episodes & Trial 1 & Trial 2 & Trial 3 & Trial 4 & Trial 5 & Mean & Standard Deviation \\
\hline
\multirow{2}{*}{Lunar Lander} & DS-DQN & 10000 & 1674 & 2163.8 & 1807.0 & 3322.2 & 2894.5 & \textcolor{blue}{2372.3} & 636.50 \\
 & Vanilla DQN & 10000 &  9031 & 3160.8 & 5136.7 & 2584.1 & 3098.8 & \textcolor{red}{4602.28} & 2379.52 \\
\hline
\multirow{2}{*}{Mountain Car} & DS-DQN & 5000 & 53.1 & 595.1 & 457.6 & 3259.9 & 1257 & \textcolor{blue}{1124.54} & 1135.82 \\
& Vanilla DQN & 5000 & 3049 & 78.8 & 156.8 & 5000 & 73 & \textcolor{red}{1671.52} & 2018.05 \\
\hline
\multirow{2}{*}{Breakout}  & DS-DQN & 10000 & 2383.9 & 1820.1 & 1439.6 & 2486 & 2203 & \textcolor{blue}{2066.52} & 387.06 \\
& Vanilla DQN & 10000 & 3415 & 2711.3 & 2416.8 & 2325.2 & 1329.9 & \textcolor{red}{2439.64} & 673.83 \\
\hline
\multirow{2}{*}{Space Invaders} & DS-DQN & 10000 & 1221.7 & 2266.4 & 1550.5 & 2887.4 & 1449.6 & \textcolor{blue}{1875.12} & 615.01 \\
& Vanilla DQN & 10000 & 4699.4 & 3243.6 & 2614.8 & 10000 & 9524.9 & \textcolor{red}{6016.54} & 3136.00 \\
\hline
\multirow{2}{*}{Asterix} & DS-DQN & 30000 & 6767.4 & 3557.3 & 3935.8 & 3837.1 & 6298.2 & \textcolor{blue}{4879.16} & 1363.98 \\
& Vanilla DQN & 30000 & 30000&	30000&	30000	& 30000&	29749.9&	\textcolor{red}{29949.98} & 100.4 \\
\hline
\end{tabular}

\end{table*}

We implement our proposed ``DS-DQN'' (Algorithm \ref{UC-DQN}) on all five tasks: Mountain Car, Lunar Lander, and the three MinAtar games. For comparison, we also implement the vanilla DQN algorithm \cite{dqn}. The objective of our work is to demonstrate that adding a simple regularization term to the DQN loss function can generate richer and more distinguishable state representations compared to standard DQN. This directly addresses a known limitation of DQN, where overgeneralization can hinder the transferability of learned features.
Given this focus, we chose vanilla DQN as the baseline for our experiments, as our proposed DS-DQN builds directly upon it by introducing the correlation-penalizing term. 
We ensure that both algorithms use identical neural network architectures and hyperparameters for a fair evaluation.
After sufficiently training both algorithms, we extract the learned state representations $\phi(s)$ from each and then apply linear function approximation (LFA) Q-learning across all tasks.
For Mountain Car and Lunar Lander, we test transfer learning by applying LFA Q-learning to new tasks, where the goal states are perturbed with zero mean Gaussian noise to simulate small variations in the task. For the MinAtar games, each time the environment is initialized, the stochastic nature of the environment introduces enough variability to treat each initialization as a new task. This allows us to evaluate how well the learned state representations could be transferred in a more dynamic and unpredictable setting.

Note that in Step 12 of the Algorithm \ref{UC-DQN}, it is necessary to sample $m$ state pairs to calculate the empirical correlation coefficient. We utilize a heuristic that leverages the structure of the tasks considered in the experiments, rather than relying on random sampling. Specifically, we first apply k-means clustering to the raw state representations of the tasks, grouping the states into $k$ clusters. To compute the correlation coefficient, we then sample one state from each cluster and calculate the average correlation across these $k$ pairs. This approach exploits the inherent structure of the tasks, helping to enforce distinguishability in the state representations more effectively. 
We note that random sampling approach in Algorithm \ref{UC-DQN} is the default and general-purpose approach, while k-means was used as an experimental heuristic in this specific context.

The objective of our experiments is to address two key questions: \begin{enumerate} \item How does DS-DQN perform during training compared to vanilla DQN? \item How well does Linear Function Approximation (LFA) Q-learning perform when applied to the state representations transferred from the trained DS-DQN, compared to vanilla DQN? \end{enumerate}
To address the first question, we present plots showing the moving average reward (calculated as the average return over the last $10$ episodes) across training episodes for both DQN and DS-DQN. These plots provide a comparison of the performance of both algorithms during training. 

For the second, and more important question, we evaluate the performance of LFA Q-learning when applied to the state representations learned by 
each algorithm. Specifically, we take the best-performing network configuration from each algorithm after training and use the corresponding state representations for LFA Q-learning. To assess the transferability of these representations, we run LFA Q-learning until predefined exit conditions are met for each task \footnote{The exit conditions, hyperparameters, and complete source code are provided in the following GitHub repository: \href{https://anonymous.4open.science/r/ICTAI_211-07B5/}{link}.} 

\textbf{Note on Transfer Results:} RL algorithms are known to exhibit significant stochasticity, arising from factors such as random initial states and the sampling of state-action pairs. Given this variability, it is crucial to run both the DS-DQN and LFA experiments multiple times to ensure statistically reliable results.
To achieve this, we employ the following methodology: First, we run the DS-DQN algorithm and transfer the learned state representations to LFA, recording the number of episodes required for LFA Q-learning to meet the predefined exit conditions. The LFA Q-learning process is repeated 10 times, and we report the average number of episodes across these runs.
Furthermore, this entire procedure, i.e., training DS-DQN and running LFA Q-learning $10$ times, is repeated $5$ times, which we refer to as a \textit{trial}. Therefore, in total, we train DS-DQN five times and run LFA Q-learning ten times for each trained model, effectively presenting results averaged across $50$ random runs (Table \ref{main_table}). The same experimental procedure is followed for the vanilla DQN as well, ensuring a fair and comprehensive comparison.

\subsection{Discussion of Results}
In Figures \ref{rew_episode} and \ref{rew_episode_2}, we present the performance of our proposed DS-DQN and the vanilla DQN during training. As shown, the performance of DS-DQN is comparable to that of the DQN (not consistently faster or slower). This is expected as the additional regularization term aimed at reducing positive correlation between state representations can, at times, slow the learning. The true advantage of DS-DQN becomes apparent when we transfer the learned representations to LFA Q-learning, which we discuss below. 

In Table \ref{main_table}, we present the performance of LFA Q-learning using the state representations obtained from both the trained DS-DQN and vanilla DQN. The results show that our proposed DS-DQN significantly outperforms vanilla DQN, as the average number of episodes required to reach the exit condition is considerably lower for DS-DQN. In fact, vanilla DQN often struggles to meet the exit conditions within the allowed maximum number of episodes, highlighting the difficulty it faces in transferring learned representations effectively. 

Moreover, the performance gap is even more pronounced in the MinAtar games, suggesting that our algorithm excels in more complex and stochastic environments. The results in Table \ref{main_table} are particularly significant because they demonstrate that, given compact and well-structured state representations, LFA Q-learning is capable of training on challenging environments such as the Atari games. To the best of our knowledge, this is the first instance where successful training of complex games like Atari has been demonstrated through an LFA framework.
\subsection{Ablation studies}
We conduct the following ablation studies to further investigate and analyze the performance of our proposed method.
\subsubsection{LFA Q-Learning using Raw State Features} 
While we successfully transferred the learned state representations to LFA Q-learning, we now assess how significant this result truly is. To do so, we run LFA Q-learning directly on the raw state features of the environment, without using the learned representations. In Table \ref{vanilla_q}, we report the performance on all five tasks. We notice that LFA Q-learning was unable to meet the exit conditions within stipulated maximum episodes, further underscoring the effectiveness of our proposed approach.

\subsubsection{Impact of Number of Clusters and maximum Regularization Coefficient}
Two key hyperparameters in our proposed algorithm are the number of clusters ($k$) and the maximum regularization coefficient ($\lambda_{max}$). In Table \ref{lambda}, we report the results of various combinations of these parameters on the Lunar Lander task. As shown, not all parameter combinations yield good results, highlighting the importance of careful tuning. Specifically, with a lower number of clusters ($k = 10$), we observed high variance in the results, with some trials showing no convergence (e.g., ``DS-DQN'' failed to exit in the second trial for $\lambda_{max} = 0.05$). On the other hand, increasing the number of clusters to $k = 50$ resulted in significantly longer computation times. A value of $k = 25$ provided a good balance, achieving optimal convergence while maintaining reasonable training times and therefore we fix this value in our main experiments.
\begin{table}[ht]
\centering
    \caption{LFA Q-learning on raw input states}
    \label{vanilla_q}   
    \begin{tabular}
    {|l|c|c|c|}
    \hline
     Environment & Trial 1 & Trial 2 & Trial 3 \\
    \hline
    LunarLander & 10000 & 10000 & 10000 \\
    Mountain Car & 5000 &5000 & 5000\\
    Breakout & 10000 & 10000 & 10000 \\
    Space Invaders & 10000 & 10000 & 10000 \\
    Asterix & 30000 & 30000 & 30000 \\
    \hline
    
    \end{tabular}
    \end{table}
    \begin{table}[ht]
    \centering
    \caption{Performance comparison across different values of $\lambda_{max}$ and $k$ on Lunar Lander} 
    \label{lambda}
    \begin{tabular}
    {|c|c|c|c|c|}
    \hline
    $\lambda_{max}$ & $k$ & Trial 1 & Trial 2 & Trial 3 \\
    \hline
    0.001 & 25 & 3062.1 $\pm$ 811.7 & 1772.2
$\pm$ 389.1 & 5335.2 $\pm$ 4261.2 \\
    0.1 & 25 & 2351 $\pm$ 1087.0 & 1292.3 $\pm$ 381.7 & 1815 $\pm$ 292.9\\
    0.01 & 25 & 1245.6 $\pm$ 358.1 & 2739.5 $\pm$ 253.0 & 5412.5 $\pm$ 2099.6\\
    0.05 & 10 & 1098.3 $\pm$ 443.2 & - & 10000 $\pm$ 0 \\
    0.01 & 25 & 1245.6 $\pm$ 358.1 & 1028.8 $\pm$ 163.8 & 2739.5 $\pm$ 253.0 \\
    0.01 & 50 & 2473 $\pm$ 509.0 & 2264.6 $\pm$ 620.3 & 1792.6 $\pm$ 107.7\\
    \hline
    \end{tabular}
    \end{table}
\subsubsection{Correlation Between States at Exit}
\begin{figure}[ht]
    \centering
    \includegraphics[width=0.35\textwidth]{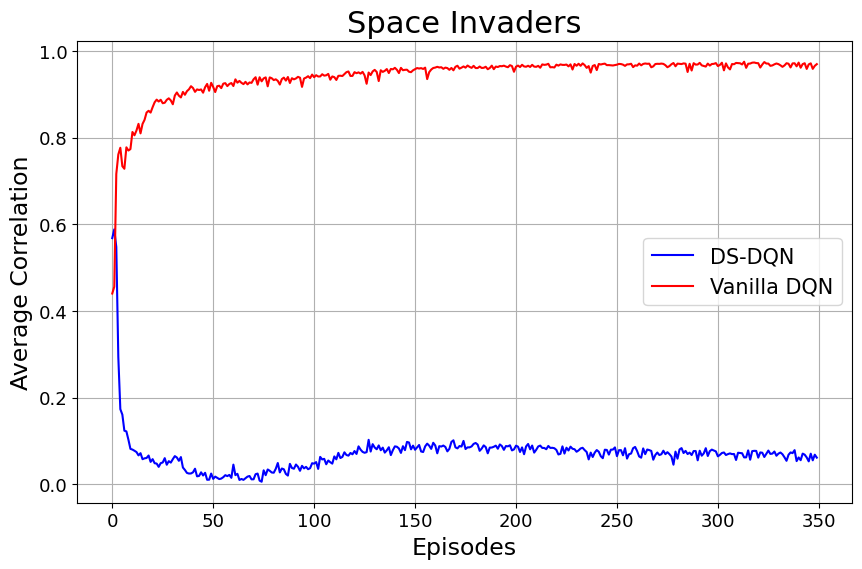} 
    \caption{Average Correlation values across episodes for the Space Invaders Environment}
    \label{corr_space}
\end{figure}

In this experiment, we analyze the average correlation between state representations learned by our proposed DS-DQN and the vanilla DQN. As shown in Table \ref{corr_vals}, the average correlation of the state representations produced by DS-DQN is significantly lower than that of vanilla DQN. Figure \ref{corr_space} illustrates how the average correlation evolves over training episodes in the Space Invaders environment. With the additional regularization term, DS-DQN shows a reduction in average correlation over the course of training, whereas, for vanilla DQN, the average correlation tends to increase, demonstrating the over-generalization problem. As shown in Table \ref{main_table}, this over-generalization severely limits the transferability of representations to LFA. In contrast, DS-DQN effectively produces distinguishable state features, enabling successful transfer to LFA.

\begin{table}[ht]
\centering
    \caption{Correlation values at the end of training}
    \label{corr_vals}   
    \begin{tabular}
    {|l|c|c|c|}
    \hline
     Environment & DS-DQN & Vanilla DQN  \\
    \hline
    LunarLander & 0.155 & 0.536  \\
    Mountain Car  & 0.935 & 0.999 \\
    Breakout & 0.0276 & 0.4251 \\
    Space Invaders & 0.0844 & 0.9531 \\
    Asterix & -0.0162 & 0.9565 \\ 
    \hline
    
    \end{tabular}
    \end{table}

\section{Conclusion}
In this work, we propose a novel DQN architecture that simultaneously computes the optimal policy and generates distinguishable state representations. This is accomplished by incorporating a regularization term that penalizes high positive correlation between the representations of different states.
We then leverage these learned representations to train new tasks using computationally simpler Linear Function Approximation Q-learning. The effectiveness of our approach is demonstrated across five non-trivial RL tasks, highlighting the benefits of the proposed scheme for efficient transfer learning. Additionally, we conduct ablation studies to analyze the impact of various components of the model.





\bibliographystyle{plain} 
\bibliography{sample}


\end{document}